\RequirePackage{fix-cm}
\documentclass[twocolumn]{svjour3}       
\smartqed  

\usepackage[table,xcdraw]{xcolor}
\usepackage{tabularx, booktabs}
\usepackage{graphicx}
\usepackage{amsmath,amssymb} 
\usepackage[ruled]{algorithm2e}
\usepackage{subfig}
\usepackage[noadjust]{cite}
\usepackage{hhline, float}
\usepackage{mathtools}
\usepackage{colortbl}

\DeclareMathOperator*{\argmin}{argmin}

\newcommand{\norm}[1]{\lVert#1\rVert}

\begin{document}

\title{Real-time Crowd Tracking using Parameter Optimized Mixture of Motion Models}


\author{Aniket Bera         \and
        David Wolinski \and
				Julien Pettr\'{e} \and
				Dinesh Manocha 
}


\institute{Aniket Bera and Dinesh Manocha \at
           Department of Computer Science \\
					The University of North Carolina at Chapel Hill\\
					NC 27599, USA\\
              \email{\{ab, dm\} @cs.unc.edu}           
           \and
           David Wolinski and Julien Pettr\'{e} \at
           INRIA-Rennes, Campus de Beaulieu\\
					 35042 Rennes Cedex, FRANCE\\
					\email{\{julien.pettre, david.wolinski\} @inria.fr}
}

\date{Received: date / Accepted: date}

\maketitle

\begin{abstract}

We present a novel, real-time algorithm to track the trajectory of each pedestrian in moderately dense crowded scenes.
Our formulation is based on an adaptive particle-filtering scheme that uses a combination of various multi-agent heterogeneous pedestrian simulation models.
We automatically compute the optimal parameters for each of these different models based on prior tracked data and use the best model as motion prior for our particle-filter based tracking algorithm.
We also use our ``mixture of motion models'' for adaptive particle selection and accelerate the performance of the online tracking algorithm.
The motion model parameter estimation is formulated as an optimization problem, and we use an approach that solves this combinatorial optimization problem in a model independent manner and hence scalable to any multi-agent pedestrian motion model.
We evaluate the performance of our approach on different crowd video datasets and highlight the improvement in accuracy over homogeneous motion models and a baseline mean-shift based tracker. 
In practice, our formulation can compute trajectories of tens of pedestrians on a multi-core desktop CPU in in real time and offer higher accuracy as compared to prior real time pedestrian tracking algorithms. 

\end{abstract}


\section{Introduction}\label{section:introduction}

The tracking of human crowd motion is becoming increasingly ubiquitous.
It is a well-studied problem that has many applications in surveillance, behavior modeling, activity recognition, disaster prevention, and the analysis of crowd phenomena.
Despite many recent advances, it is still difficult to accurately track pedestrians in real-world
scenarios, especially as the crowd density increases.
The problem of tracking pedestrians and objects has been studied in computer vision and image processing for three decades.
However, tracking pedestrians in a crowded scene is regarded as a hard problem due to the following reasons: intra-pedestrian occlusion (one pedestrian blocking another), changes in lighting and pedestrian appearance, and the difficulty of modeling human behavior or the intent of each pedestrian. 

\begin{figure*}[ht]
	\centering
		\includegraphics[width=1.0\textwidth]{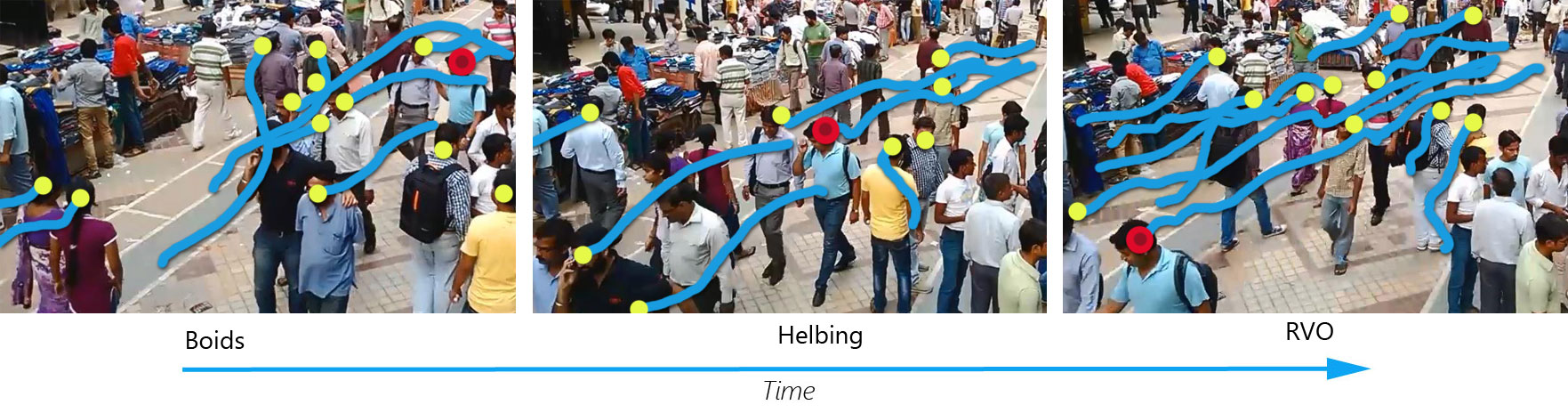}
\caption{{
Our mixture motion model can accurately compute the trajectories in real time.
We highlight different motion models (Boids, Helbing's Social Forces, or RVO) used for the same pedestrian (marked in red) over different frames. We adaptively choose the best-fit model for every pedestrian in the scene.
This increases the accuracy by 4\%-18\% of our adaptive tracking algorithm.
}}
\label{fig:1}
\end{figure*}

One approach that improves the accuracy of tracking algorithms is the use of realistic crowd motion models.
These motion models simulate the current behavior of each pedestrian in the crowd in order to predict the pedestrians' possible future positions.
There has been considerable work on developing crowd motion models for pedestrians in the areas of computer graphics, robotics, computer animation, and pedestrian dynamics. Many approaches have been investigated that suggest different \textit{principles} to model crowds.Most of these models use kind of parameters to describe the shape and trajectory of each agent. While many approaches have been investigate to model the motion of the agents, there is relatively less effort to estimating model parameters based on available data, evaluating and comparing the effects of these parameters, and quantifying the improvements that can result from parameter optimization.

Prior realtime or online crowd-tracking algorithms use a single, homogeneous motion model.
Every motion model is unique and generally relies upon one or more assumptions: these include the assumption of highly coherent motion in terms of velocity or acceleration, or assumptions about how pedestrian trajectories will change in response to other agents or obstacles.

The simpler motion models assume that agents will ignore any interactions with other pedestrians, instead assuming that they will follow ``constant-speed'' or ``constant-acceleration'' paths to their immediate destinations.
However, the accuracy of this assumption decreases as crowd density in the environment increases (e.g. to 2-4 pedestrians per square meter).
More sophisticated pedestrian motion models take into account interactions between pedestrians, formulated either in terms of attraction or repulsion forces or collision-avoidance constraints.

In real-world scenarios, the trajectory of each pedestrian is governed by its intermediate goal location, intrinsic behaviors, as well as local interactions with other pedestrians and obstacles in the scene.
In a dense crowd setting, the behavior of each pedestrian changes in response to the environment, the overall crowd density and flow, and the behavior of other pedestrians.
It may not be possible, therefore, to model the overall behavior of each pedestrian with a single, homogeneous motion model.
Furthermore, each of these homogeneous models is described using some parameters that may correspond to the size, speed, anticipation period, or local navigation constraints of each pedestrian.
The accuracy of each motion model is governed by the choice of these parameters.
As the behavior of each pedestrian responds to changes in a dynamic environment, these model parameters should be recomputed or updated to improve the resulting motion model's accuracy.
Overall, we need efficient techniques that can take into account heterogeneous behaviors based on constantly changing models and underlying parameters.
\\\\

\begin{figure*}[ht]
	\centering
		\includegraphics[width=1.0\textwidth]{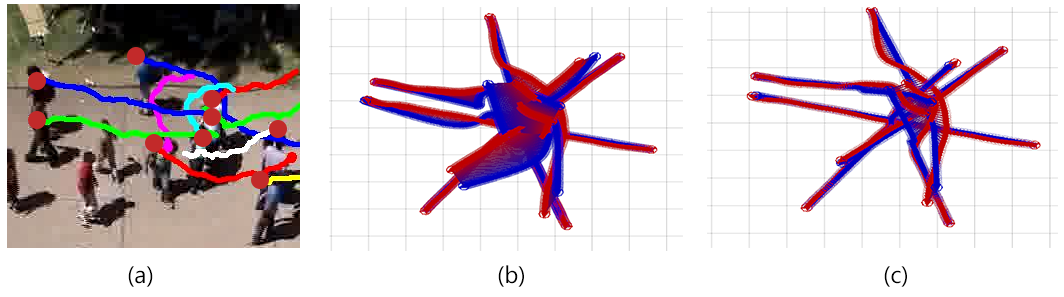}
\caption{{
(a) Our tracking on the \textit{student003} dataset~\cite{student003}.
(b) Comparing ground truth (in red) with with prediction by Social Forces by Helbing et al.~\cite{helbing1995social}(in blue).
(c) Comparing ground truth (in red) with with prediction by our motion model mixture (in blue).
The distance between the red and blue points denote the error in prediction. We can see that the error with our approach is considerably lower.
}}
\label{fig:1}
\end{figure*}

{\bf Main results:}
We present a method that uses particle filters to perform realtime pedestrian tracking in moderately crowded scenes.
Our formulation computes the best-fit or mixture motion model for each pedestrian based on prior tracked data.
In order to characterize the heterogeneous, dynamic behavior of each agent, we use an optimization based scheme to perform the following steps:
\begin{itemize}
\item Choose, every few frames, the new motion model that best describes the local behavior of each pedestrian based on tracked data.
\item Compute the optimal set of parameters for that motion model that best fit this tracked data.
\item Computing the adaptive number of particles for each pedestrian based on a combination of metrics for optimizing performance.
\end{itemize}
We compute the locally-optimal motion model for the pedestrians in realtime and use that, along with a particle-filter based tracker, to compute their trajectories.
In our approach, we consider a variety of possible motion models to characterize pedestrian motion during each frame: Boids~\cite{Reynolds1987}, Social Forces~\cite{helbing1995social} or reciprocal velocity obstacles~\cite{van2011reciprocal} as possible models to characterize the motion of a pedestrian during each frame. For videos with high fps(over ~50 fps), a constant velocity model may be sufficient to model the motion prior.
Furthermore, we use our heterogeneous motion model to adaptively choose the number of particles for each pedestrian in our particle-filter.
This adaptive formulation can increase the runtime speed based of our system based on a reliability measure computed using mixture motion model.
We evaluate our method in comparison with homogeneous motion models on high definition crowd datasets that include both indoor and outdoor scenes recorded at different locations with 30 - 150 pedestrians and also standard datasets used in the pedestrian tracking community.

In practice, our adaptive particle-filter tracker with adaptive motion model is about 4-18\% more accurate than prior interactive tracking algorithms that use homogeneous or simple motion models.
Moreover, as the crowd density increases, we observe increased improvements in the level of accuracy.
Moreover, the adaptive particle selection can increase the runtime frame rate by 2-2.5 times as compared to algorithms that use a constant high number of particles.
Overall, our algorithm can track tens of pedestrians at realtime rates (i.e. more than 25fps) on a multi-core CPU.

The rest of the paper is organized as follows.
In Section~\ref{section:related_work}, we give an overview of prior work related to online pedestrian tracking.
Section~\ref{section:overview} gives an overview of our approach, and Section~\ref{section:mixture_motion_model} describes our multi-agent heterogeneous motion model.
Section~\ref{section:results} evaluates the different components of our algorithm and compares it with other online tracking methods.

\section{Related work}\label{section:related_work}

In this section, we briefly review some prior work on pedestrian tracking and motion models.
Multi-pedestrian tracking has attracted a lot of research attention in recent years.
We refer the reader to some excellent surveys~\cite{wuonline, enzweiler2009monocular,yilmaz2006object}.

At a broad level, pedestrian tracking algorithms can be classified as either online or offline trackers.
Online trackers use only the present or previous frames for realtime tracking. 
Zhang et al.~\cite{zhang2012real} proposed an approach that uses non-adaptive random projections to model the structure of the image feature space of objects, and Tyagi et al.~\cite{tyagi2008context} described a technique to track pedestrians using multiple cameras.
Offline trackers, on the other hand, use data from future frames as well as current and past data~\cite{sharma2012unsupervised, rodriguez2011density}.
These methods, however, require future-state information; they are therefore not useful for realtime applications.

In addition to the online/offline classifications, tracking algorithms can also be classified based on their underlying search mechanisms: as either deterministic or probabilistic trackers. 
Deterministic trackers iteratively attempt to search for the local maxima of a similarity measure between the target candidate (the location of the pedestrian in a frame) and the object model (the initial state of the pedestrian).
The most commonly used deterministic trackers are the mean-shift algorithm~\cite{yilmaz2007object} and the Kanade-Lucas-Tomasi algorithm~\cite{lucas1981iterative}.
In probabilistic trackers, the movement of the object is modeled based on its underlying dynamics.
Two well-known probabilistic trackers are the Kalman filter and the particle filter.
Particle filters are more frequently used than Kalman filters in pedestrian tracking, since particle filters are multi-modal and can represent any shape using a discrete probability distribution.
 
\emph{Motion Models:}
The problem of modeling crowd behaviors and motions has received significant attention in various disciplines.
This attention has resulted in a high number of simulation models based on microscopic or macroscopic principles. 
Several of the proposed motion models represent each individual or pedestrian in a crowd as particles (or as 2D circles in a plane), then model the interactions between these particles.
Reynolds'~\cite{Reynolds1987, Reynolds1999} seminal approach is representative of such models: local interactions, matching an agent's speed and orientation to those of its neighbors, determine agents' motions and lead to emergent behaviors.
Many popular algorithms model agents as particles which are subjected to repulsive forces~\cite{helbing1995social} and additional behavior-improving rules.
More recently, velocity-based algorithms~\cite{van2011reciprocal, Pettre2009, Karamouzas2009} have been developed, which model agents' motions in velocity-space to ensure collision-free trajectories over short future time windows.
Other approaches that have recently been developed are based on cognitive models~\cite{chung2010mobile}, affordance~\cite{Fajen2007}, short-term planning using a discrete approach~\cite{Antonini2006} or Linear Trajectory Avoidance (LTA)~\cite{pellegrini2009you}.
A final recent approach uses the virtual optic flow of agents to derive perceptual variables in order to compute collision-free motions~\cite{Ondvrej2010}.
A few tracking algorithms use the Reciprocal Velocity Obstacle (RVO) model as motion prior~\cite{Liu2014}~\cite{bera2014}.

Many non-particle-based motion modeling techniques have also been proposed; these techniques are useful mainly for crowded scenes in which pedestrians display similar motion patterns.
Song et al.~\cite{song2013fully} proposed an approach that clusters pedestrian trajectories based on the assumption that ``persons only appear/disappear at entry/exit.''
Ali et al.~\cite{ali2008floor} presented a floor-field based method to determine the probability of motion in densely crowded scenes. 
Rodriguez et al.~\cite{rodriguez2011data} used a large collection of public crowd videos and learned about crowd motion patterns by extracting global video features.
Kratz et al.~\cite{kratz2012going} and Zhao et al.~\cite{zhao2012tracking} used local motion patterns in dense videos for pedestrian tracking.
Shu et al.~\cite{kratz2012going} proposed an approach that learns part-based person-specific SVM classifiers which capture dynamically changing pedestrian appearance. Zamri et al.~\cite{kratz2012going} used generalized minimum clique graphs for multiple-person tracking. Leal-Taix\'{e} et al.~\cite{leal2012exploiting} used a social and grouping behavior as a physical model in their tracking system. Burgos-Artizzu et al.~\cite{burgos2012social} presented a novel method for analyzing social behavior, particularly in mice videos, where the continuous videos are segmented into action `bouts' by building a temporal context model.

These methods are well-suited for modeling motion in dense crowds with few distinct motion patterns; however, they may not work in heterogeneous crowds.


\section{Our Approach}\label{section:overview}

In this section, we give an overview of our approach.
First, offer an overview of our method, which is followed by more detailed explanations of the various components of our realtime tracking algorithm.

\subsection{Overview}



\begin{figure*}
	\centering
\includegraphics[width=1.0\textwidth]{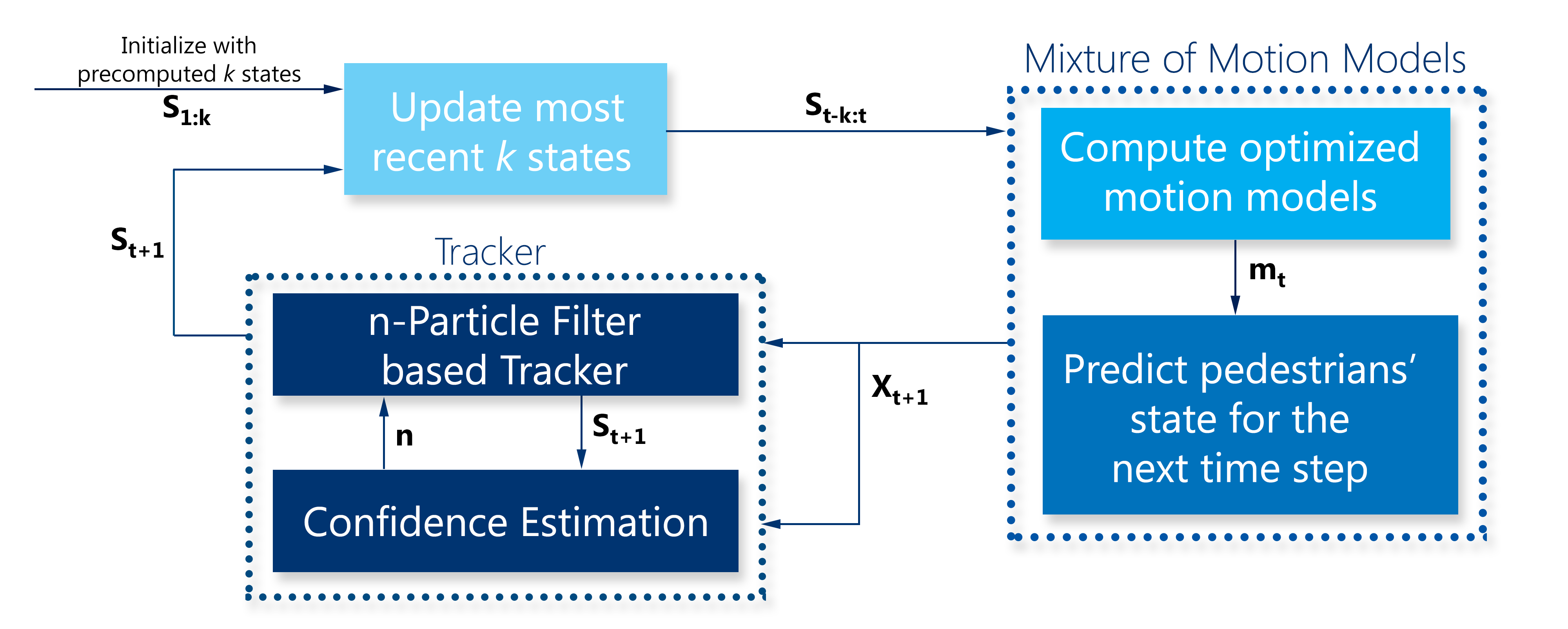}
\caption{
Overview of our realtime tracking algorithm.
The symbols used in this figure are explained in Section~\ref{sec:notations}.
We use the trajectory computed over prior k frames, expressed as a succession of states, to compute the new motion model; we use our mixture motion model to compute next states using a particle filter.
}
	\label{fig:unified}
\end{figure*}

Our approach can be viewed as a feedback pipeline (Figure~\ref{fig:unified}). We use the most recent states (positions and velocities) for each agent and use them to compute our mixture model. This mixture model is used to predict next state of the pedestrian for the next frame.
In other words, the next state is used as motion prior input for the tracker; it is also combined with confidence estimation computation to dynamically compute the number of particles.
As a final step, the tracker's definitively estimated next state is fed back into the loop, becoming the most recent agent state.

\textbf{Data Representation} Our algorithm keeps track of the {\em state} (i.e. position and velocity) of each pedestrian for the last k timesteps or frames.
These are referred to as the k-states of each pedestrian.
%
These k-states are initialized by pre-computing the states from the first k timesteps.
The k-states are updated at each timestep by removing the agents' state from the oldest frame and adding the latest tracker-estimated state.

\textbf{The mixture motion model} is a combination of several independent motion models that are widely used for pedestrian modeling in crowds: Boids, RVO and Social forces.
This mixture motion model is used to compute the best motion model for the agents during each frame.
First, based on an optimization algorithm, we ``configure'' (see Section~\ref{section:mixture_motion_model}) the motion models to ``best'' match the recent k-states data and select the best model based on a specific metric.
Second, we use the ``best configured'' motion model to make a prediction on the agents' next state.

\textbf{The tracker} is a particle-filter based tracker which uses the motion prior, obtained from the mixture of motion models, to estimate the agents' next state.
This tracker further uses a confidence estimation stage to dynamically compute the number of particles that balance the tradeoffs between the computation cost and the accuracy.



\subsection{Notation and Terminology}\label{sec:notations}

We use the following notations in our paper:

\begin{itemize}
 \item $S$ represents the state (position and velocity) of an arbitrary pedestrian as computed by the tracker
 \item $X$ represents the state (position and velocity) of an arbitrary pedestrian inside a crowd motion model
 \item $m$ represents the ``best configured'' motion model from the mixture of motion models $\{ f1, f2, ... \}$
 \item bold fonts are used to represent values for all the pedestrians in the crowd; for example $\mathbf{S}$ represents the states (positions and velocities) of all pedestrians as computed by the tracker
 \item subscripts are used to indicate time; for example $m_t$ represents the ``best configured'' motion model at timestep $t$, and $\mathbf{S}_{t-k:t}$ represents all states of all agents for all successive timesteps between $t-k$ and $t$, as computed by the tracker.
\end{itemize}

The ``best configured'' motion model can then be used as follows: $X_{t+1} = m_t(X_t)$ or $\mathbf{X}_{t+1} = m_t(\mathbf{X}_t)$ to compute the motion of one arbitrary pedestrian or all pedestrians, respectively.


\subsection{Particle Filter for Tracking}

Though any online tracker which requires a motion prior system can be used, we use particle filters as the underlying tracker algorithm.
The particle filter is a parametric method which solves non-Gaussian and non-linear state estimation problems~\cite{arulampalam2002tutorial}.
Particle filters are frequently used in object tracking, since they can recover from lost tracks and occlusions.
The particle tracker's tracking uncertainty is represented in a Markovian manner by only considering information from present and past frames.

%
Here, we consider the ``best configured'' motion model $m_t$ as well as the error $Q_t$ in the prediction that this ``best configured'' motion model generated.
Additionally, the observations of our tracker can be represented by a function $h()$ that projects the state $X_{t}$ to a previously computed state $S_{t}$.
Moreover, we denote the error between the observed states and the ground truth as $R_t$.
We can now phrase them formally in terms of a standard particle filter as below:

\begin{equation}
S_{t+1} = m_t(X_{t}) + Q_t,
\end{equation}
\begin{equation}
S_{t} = h(X_{t}) + R_t.
\end{equation}

Particle filtering is a Monte Carlo approximation to the optimal Bayesian filter, which monitors the posterior probability of a first-order Markov process:

\begin{equation}
\begin{multlined}
p(X_{t} | S_{t-k:t}) = \\
\alpha p(S_{t} | X_{t}) \int_{X_{t-1}}{p(X_{t}|X_{t-1})p(X_{t-1}|S_{t-k:t-1})}dX_{t-1},
\end{multlined}
\end{equation}

\noindent where $X_{t}$ is the process state at time $t$, $S_{t}$ is the observation, $S_{t-k:t}$ is all of the observations through time $t$, $p(X_{t}|X_{t-1})$ is the process dynamical distribution, $p(S_{t}, X_{t})$ is the observation likelihood distribution, and $\alpha$ is the normalization factor.
Since the integral does not have a closed form solution in most cases, particle filtering approximates the integration using a set of weighted samples ${X_{t}^{(i)},\pi_{t}^{(i)}}_{i=1,...,n}$, where $X_{t}^{(i)}$ is an instantiation of the process state, known as a particle, and $\pi_{t}^{(i)}$'s are the corresponding particle weights.
With this representation, the Monte Carlo approximation to the Bayesian filtering equation is:

\begin{equation}
p(X_{t} | S_{t-k:t}) \approx \alpha p(S_{t} | X_{t}) \sum_{i=1}^{n}{\pi_{t-1}^{(i)}p(X_{t}^{(i)}) | p(X_{t-1}^{(i)})},
\end{equation}

\noindent where $n$ refers to the number of particles.

In our formulation, we use the motion model to infer dynamic transition, $p(X_{t}|X_{t-1})$, for particle filtering.


We optimize our computation speed by adaptively modifying the number of active particles in our system using a combination of confidence metrics. A brief overview is given in Section~\ref{sec:adaptive_particle_selection}.

\section{Mixture Motion Model}\label{section:mixture_motion_model}

In this section, we introduce the notion of a parameterized motion model.
We then describe the different parameterized motion models that form the basis for the mixture motion model.
Finally, we describe the mixture motion model itself.

\subsection{Parameterized Motion Model}

A motion model is defined as an algorithm $f$ which, from a collection of agent states $\mathbf{X}_t$, derives new states $\mathbf{X}_{t+1}$ for these agents, representing their motion over a timestep towards the agents' immediate goals $\mathbf{G}$:
\begin{align}
\mathbf{X}_{t+1} = f(\mathbf{X}_t,\mathbf{G}).
\end{align}

Motion algorithms usually have several parameters that can be
tuned in order to change the agents' behaviors.
We assume that each parameter can have a different value for each pedestrian.
By changing the value of these parameters, we get some variation in the resulting trajectory prediction algorithm. 
We use $\mathbf{P}$ to denote all the parameters of all the pedestrians.
Typically, for a crowd of 50 pedestrians, the dimension of $\mathbf{P}$ could be anywhere in the range 150-300 depending on the motion model. 
In our formulation, we denote the resulting parameterized motion model as:
\begin{align}
\mathbf{X}_{t+1} = f(\mathbf{X}_t,\mathbf{G},\mathbf{P}).
\label{eqn:crowdSim}
\end{align}

\subsection{Motion Models}

Our mixture motion model can include any generic motion model that conforms to Equation~(\ref{eqn:crowdSim}).
Here we describe the three component motion models that currently make up the mixture motion model in our current implementation.

\subsubsection{Reciprocal Velocity Obstacles}

RVO is a local collision-avoidance and navigation algorithm.
Given each agent's state at a certain timestep, it computes a collision-free state for the next timestep\cite{van2011reciprocal}.
Each agent is represented as a 2D circle in the plane, and the parameters (used for optimization) for each agent consist of the representative circle's radius, maximum speed, neighbor distance, and time horizon (only future collisions within this time horizon are considered for local interactions).

Let $V_{pref}$ be the preferred velocity for a pedestrian that is based on the immediate goal location. The RVO formulation takes into account the position and velocity of each neighboring pedestrian to compute the new velocity. The velocity of the neighbors is used to formulate the ORCA constraints for local collision avoidance~\cite{van2011reciprocal}. The computation of the new velocity is expressed as an
optimization problem for each pedestrian.
If an agent's preferred velocity is forbidden by the ORCA constraints, that agent chooses the closest velocity that lies in the feasible region: 

\begin{equation} \label{eqn:ORCA}
V_{RVO} = \underset{V \notin ORCA}{\arg\max} \|V - V_{pref}\|.
\end{equation}

More details and mathematical formulations of the ORCA constraints are given in~\cite{van2011reciprocal}.
As per Equation~(\ref{eqn:crowdSim}), $f$ returns the states obtained with the admissible velocity that is closest to the preferred velocity.

\subsubsection{The Boids Model}
Initially developed to simulate the flocking behavior of birds, this model has later been extended to pedestrian motion in a crowd.
Broadly, three rules are enforced on Boids agents:

\begin{itemize}
	\item \textbf{Separation}: steer to avoid crowding local agents
	\item \textbf{Alignment}: steer towards the average heading of local agents
	\item \textbf{Cohesion}: steer to move toward the average position (center of mass) of local agents
\end{itemize}

Thus, as per Equation~(\ref{eqn:crowdSim}), $f$ is a function of agents' positions at some specified future time (current time plus constant).
When the predicted distance between the pedestrians gets too low, a separation force is computed and added to the attraction force that is pulling the agents toward their goal.
The parameters are radius (size of 2D circle agents) and comfort speed (i.e., speed when no interactions occur).

\subsubsection{Social Forces Model}
The social forces model is defined by the combination of three different forces: the personal motivation force, social forces, and physical constraints:

\begin{itemize}
	\item \textbf{Personal Motivation force}  ($F^{M}$): This is the incentive to move at a certain preferred velocity in a certain direction.

	\item \textbf{Social forces} ($F^{S}$): These are the repulsive forces from other agents and obstacles.

	\item \textbf{Physical Constraints} ($F^{P}$): These are the hard constraints other than the environment and other agents.
\end{itemize}

The net force $F^{C} = F^{M} + F^{S} + F^{P}$ then defines
an agent's chosen new velocity. For a detailed explanation of the method, refer to~\cite{helbing1995social}.

As per Equation~(\ref{eqn:crowdSim}), $f$ is a function of the agents' positions from which all computed forces are derived. The parameters are radius and comfort speed.

\subsection{Mixture of motion models}

\begin{figure*}[ht]
	\centering
\includegraphics[width=0.8\textwidth]{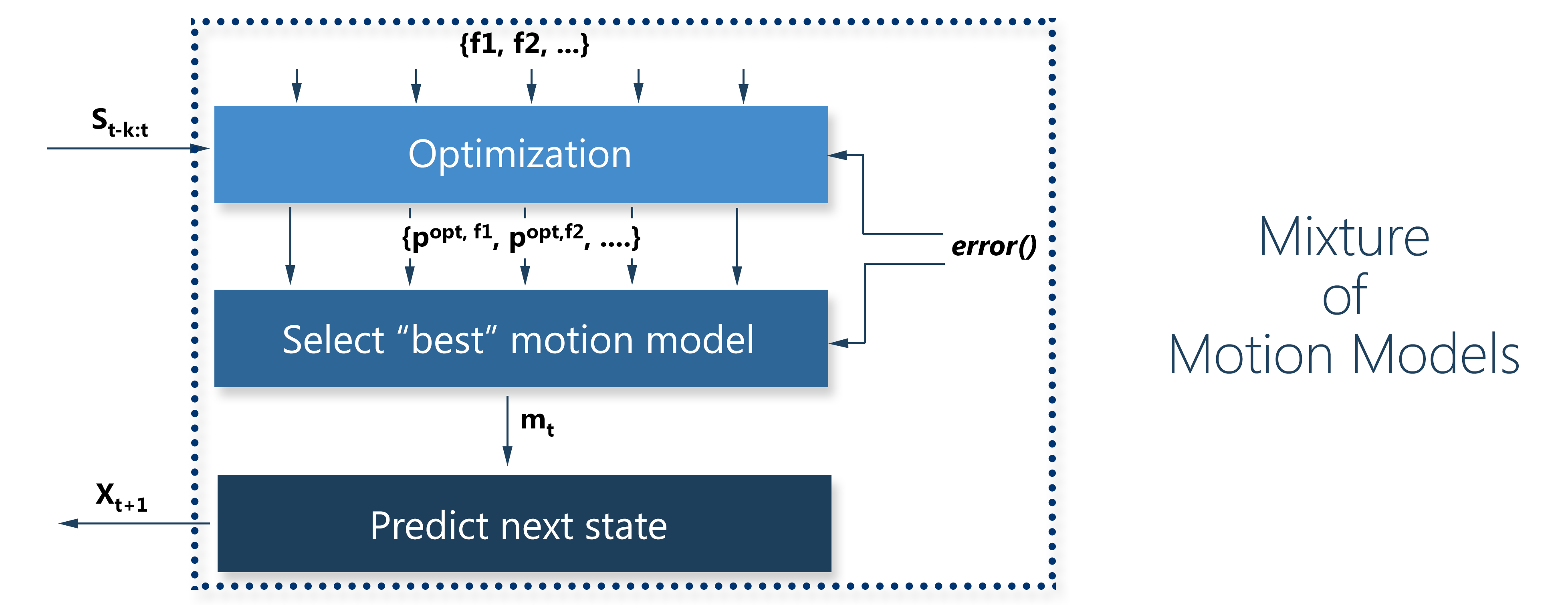}
\caption{Our parameter optimization algorithm used in Figure~\ref{fig:unified}. Based on the error metric, we compute optimal parameters for each motion model. The best motion model (from RVO, Social Forces, Boids or LIN) is used for trajectory extraction and predicting the next state.}
	\label{fig:algo}
\end{figure*}

\vspace{-0.25cm}

We now present the algorithm to compute the mixture motion model, which essentially corresponds to computing the ``best'' motion model at any given timestep.
In this case, the ``best'' motion model is the one that most accurately matches agents' immediately past states, as per a given error metric.
This ``best'' motion model is determined by an optimization framework, which automatically finds the parameters $\mathbf{P}$ that minimize the error metric. Wolinski et al. ~\cite{Wolinski2014} designed an optimization framework for evaluating crowd motion models but it computes the optimal parameters in an offline manner for a single homogenous simulation model. Our framework is online and iteratively computes the best heterogeneous motion every few frames and chooses the most optimized crowd parameters at a given time. The computation cost is considerably lower and hence useable for real-time tracking. 

\subsubsection{Formalization}

Formally, at any timestep $t$, we define the agents' (k+1)-states (as computed by the tracker) $\mathbf{S}_{t-k:t}$:
\begin{align}
\mathbf{S}_{t-k:t} = \bigcup_{i=t-k}^{t} \mathbf{S}_i.
\end{align}

Similarly, a motion model's corresponding computed agents' states $f(\mathbf{S}_{t-k:t}, \mathbf{P})$ can be defined as:
\begin{align}
f(\mathbf{S}_{t-k:t}, \mathbf{P}) = \bigcup_{i=t-k}^t f(\mathbf{X}_i, \mathbf{G}, \mathbf{P}),
\end{align}
initialized with $\mathbf{X}_{t-k} = \mathbf{S}_{t-k}$  and $\mathbf{G} = \mathbf{S}_t$.

At timestep $t$, considering the agents' k-states $\mathbf{S}_{t-k:t}$, computed states $f(\mathbf{S}_{t-k:t}, \mathbf{P})$ and a user-defined error metric $error()$, our algorithm computes:
\begin{align}
\mathbf{P}^{opt, f}_t
& = \argmin_{\mathbf{P}} error(f(\mathbf{S}_{t-k:t},\mathbf{P}),\mathbf{S}_{t-k:t}),
\label{eqn:optimize}
\end{align}
where $\mathbf{P}^{opt, f}_t$ is the parameter set which, at timestep $t$, leads to the closest match between the states computed by the motion algorithm $f$ and the agents' k-states.

For several motion algorithms $\{ f1, f2, ... \}$, we can then compute the algorithm which best matches
the agents' k-states $\mathbf{S}_{t-k:t}$ at timestep $t$:
\begin{align}
m_t = f^{opt}_t = \argmin_{f} error(f(\mathbf{S}_{t-k:t},\mathbf{P}^{opt, f}_t),\mathbf{S}_{t-k:t}),
\label{eqn:optmodel}
\end{align}
and consequently, the best (as per the error in the $error()$ metric itself) prediction for the agents' next state
obtainable from the motion algorithms for timestep $t+1$ is:
\begin{align}
\mathbf{X}_{t+1} = m_t(\mathbf{S}_t).
\label{eqn:bestPred}
\end{align}

\subsubsection{Optimization Algorithm and Error Metric}

Optimizing crowd parameters is a unique and challenging problem. Because most simulation methods have several parameters to tune for each agent, even moderately sized scenarios with a few dozen agents can become a hundred-dimensional optimization problem.

In total we tested three global optimization approaches:\textit{ Greedy algorithm}, \textit{Simulated Annealing}, and \textit{Genetic Algorithm}.

For the greedy approach we start by choosing random parameters for every agent. The chosen data similarity metric is then evaluated to establish a baseline measure of how well the simulation matches the data. After several iterations, where in each iteration starts with the best set of simulation parameter seen so far. This new set of parameters is evaluated, whichever set of parameters has the lowest error metric over all the iterations is chosen as the optimal parameters for the agents.

The main limitation with a greedy approach is that it will get stuck in local minimum in search space and also the final outcome depends on the starting point. Simulated Annealing addresses this problem. Analogous with thermodynamics, simulated annealing incorporates a `\textit{temperature}' parameter into the 
minimization procedure. At high temperatures, we explore the parameter space whereas at lower temperature, we restrict the exploration.

\begin{algorithm}
\DontPrintSemicolon
$k \leftarrow 0$\tcp*{initialize loop counter}
\While{$k<K$}{
	$T \leftarrow \operatorname{temperature}(k, K)$\tcp*{compute temperature}
	$s_{new} \leftarrow \operatorname{neighborState}(s)$\tcp*{try new neighbor}
	$e_{new} \leftarrow \operatorname{cost}(s)$\tcp*{compute cost}
	\If(\tcp*[f]{is new state better?}){$\operatorname{move}(e, e_{new}, T)$}{
		$s \leftarrow s_{new};\ e \leftarrow e_{new}$\tcp*{yes, change state}
	}
	\If(\tcp*[f]{did we find a new minimum?}){$e < e_{best}$}{
		$s_{best} \leftarrow s;\ e_{best} \leftarrow e$\tcp*{save new optimum}
		$k \leftarrow 0$\tcp*{reset loop counter}
	}
	$k \leftarrow k+1$\tcp*{increase loop counter}
}
\caption{Simulated annealing.}
\label{algo:simanneal}
\end{algorithm}

Algorithm \ref{algo:simanneal} gives the pseudocode for the process where:

\begin{description}
 \item [neighborState():] pick a new random value for a random parameter according to the parameter's base distribution
 \item [move():] is $True$ iff $e_{new}<e_{old}$, $exp({\frac{e_{old}-e_{new}}{T}})$.
 \item [temperature():] is $\frac{K-k}{K}$, $k$ being the number of iterations with no improvement and $K$ the number of such iterations allowed.
 \item [cost():] the cost as returned by the currently used metric.
\end{description}

We also use a Genetic algorithm~\cite{holland1992genetic}. The underlying optimization technique as algorithm offers the best compromise between optimization results and speed.
The efficiency component is important as our goal is realtime pedestrian tracking.

Genetic algorithms seek to overcome the problem of local minima in optimization.
This is accomplished by keeping a pool of parameter sets and, during each iteration of the optimization process, creating a new pool of potential solutions by combining and modifying these parameter sets.

 \begin{algorithm}
 \DontPrintSemicolon
 $pop \leftarrow \operatorname{initialize}()$\tcp*{initialize population}
 \While{$true$}{
 	$\operatorname{selection}(pop)$\tcp*{evaluate and select fittest}
 	\If(\tcp*[f]{should we terminate?}){$\operatorname{termination}()$}{$stop$\tcp*{yes, stop loop}}
 	$pop \leftarrow \operatorname{reproduction}(pop)$\tcp*{new generation}
 }
 \caption{Genetic algorithm.}
 \label{algo:genetic}
 \end{algorithm}
 
 Algorithm \ref{algo:genetic} provides pseudocode for the method given the following functions:

 \begin{itemize}
 	\item \textbf{initialize()}: parameters randomly initialized in accordance with the base distribution for each parameter.
 	\item \textbf{selection()}: individuals are sorted according to their score and divided into 3 groups: Best, Middle and Worst.
 	\item \textbf{termination()}: the algorithm is terminated after finding $K$ successive loop iterations without any new optimum.
 	\item \textbf{reproduction()}: based on which group it belongs to, a parameter set is attributed three probabilities $\alpha$, $\beta$ and $\gamma$. For each parameter of this individual, $\alpha$ decides if the value is changed or not, $\beta$ decides if the value is changed by crossover or mutation and, finally, $\gamma$ decides which type of mutation is done.
 	\item crossover: a crossover is done by copying a value from an individual belonging to the Best group.
 	\item mutation: a mutation is done by picking a new value at random based on either the base distribution or the current real distribution of an individual from the Best group (according to $\gamma$). 
 \end{itemize}

At each iteration, this algorithm evaluates and ranks all possible parameter sets (solutions) currently in the solution pool.
If there have been a certain number of successive iterations without any improvement, the process is terminated.
Otherwise, individual parameter values in each solution have a probability of being modified.
If so, this modification has a probability of being either a crossover or a mutation.
If it is a crossover, a value from the corresponding parameter from a better ranked solution is selected; if it is a mutation, a new value is sampled from a probability distribution.
This probability distribution can either be the one defined by the user (for instance, a preferred velocity could obey a normal law with mean $1.4 m.s^{-1}$ and standard deviation $0.3 m.s^{-1}$) or one that is computed on parameter values from better ranked solutions.
\\\\

\begin{figure}[ht]
	\centering
\includegraphics[width=0.5\textwidth]{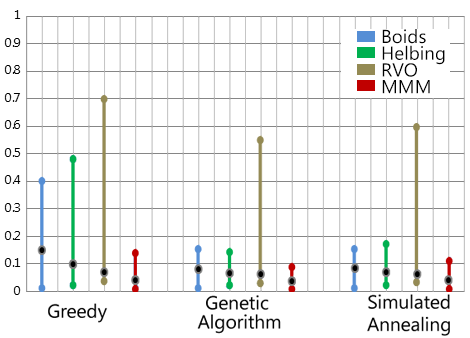}
\caption{Comparing the score of the different optimization approaches. Each graph is a range of the scores (minimum and maximum) and the black dot is the mean score. We compute the score from the normalized error metric. A lower value indicates better optimization. MMM or the `Motion-Model Mixture' is the our approach.}
	\label{fig:graph1}
\end{figure}

\begin{figure}[ht]
	\centering
\includegraphics[width=0.5\textwidth]{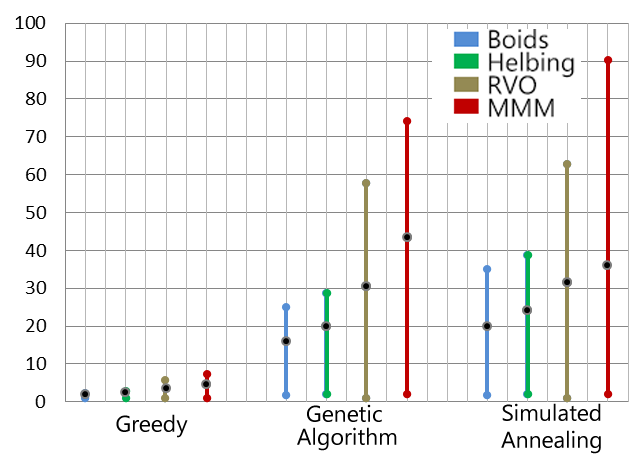}
\caption{This graph shows the time taken for each computing the every set of optimal parameters corresponding to each motion model. MMM is our approach. Time computed is in miliseconds. Each graph is a range of the scores (minimum and maximum) and the black dot is the mean score. We compute the score from the normalized error metric. }
	\label{fig:graph2}
\end{figure}

An error metric is also needed to compute the term in Equation~(\ref{eqn:optimize}).
In our case, we've chosen a metric that simply computes the average 2-norm between the observed agent positions and the tracker-computed positions.
Formally, this metric is defined at timestep $t$ as follows:
\begin{align}
error &= \sum_{i=t-k}^t \norm{\mathbf{S}_i - \mathbf{X}_i}.
\label{eqn:totaldist}
\end{align}

\begin{figure*}[ht]
	\centering
		\includegraphics[width=1.0\textwidth]{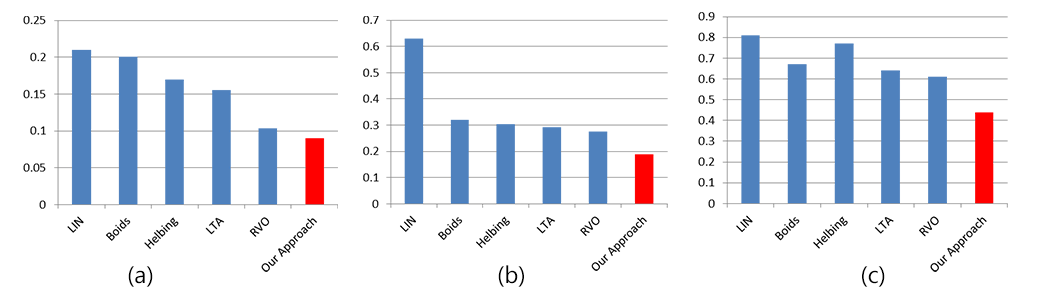}
\caption{{
This is the RMS error in the predicted position compared to the ground truth. For an unbiased comparison, all measurements are in ground-space (meters). We have divided our dataset into 3 categories (Refer table~\ref{tb:tablediv})
(a) Low-density datasets
(b) Medium-density datasets
(c) High-density datasets
We find that our approach considerably lower error for future-state prediction in medium-density crowds.
}}
\label{fig:error}
\end{figure*}

\subsection{Adaptive Particle Selection}\label{sec:adaptive_particle_selection}

The performance of a particle filter is proportional to the number of particles used for each pedestrian, and the process can be expensive for a high number of particles.
However, with more particles, the probability that a pedestrian will be tracked accurately is higher; fewer particles, though computationally less expensive, actually lowers the tracking accuracy.
As a result, we need to use an appropriate number of particles to balance the tradeoffs between computation cost and accuracy.
Ideally, one would use fewer particles most of the time, increasing their number only when needed: when there is a large change in motion trajectory, lighting, appearance or partial occlusions, for example.

To this end, we estimate tracker confidence and particle selection by using the motion model.
We analyze the confidence of our tracker given the number of particles based on combining various metrics to measure the propagation and motion model reliability.
The propagation reliability is a measure of how well the object matches the initial target candidate and also the last tracked object:

\begin{equation}
pr_{t} = g(\norm{O_{t}-O_{t-1}}, \norm{O_{t}-O_{0}}), 
\end{equation}

\noindent where $pr_{t}$ is the propagation reliability at time $t$ and $O_{t}$ denotes the object representation at time $t$. Motion model reliability is a normalized difference measure between the tracked state and the predicted state $f(X_{t-1})$ given by the motion model $f$:

\begin{equation}
mmr_{t+1} = h(\norm{f(X_{t-1}) - S_t}),
\end{equation}

\noindent where $mmr_{t+1}$ is the motion model reliability at timestep $t+1$ and $h$ is function varying linearly to the norm difference of the actual and simulated trajectories.

The combination of these metrics helps us in optimizing the number of active particles needed in the system.
In our mixture of motion models, our system chooses the optimal motion algorithm $m_t$ from all possible motion models $\{ f1, f2, ... \}$ (Equation~(\ref{eqn:optmodel})) with the optimal parameter set.
Hence the motion model reliability is always higher compared to systems with homogeneous or non-varying motion models.

\begin{equation}
mmr_{t+1}^{opt} = h(\norm{m_t(X_{t-1}) - S_t}).
\end{equation}


\begin{table}[ht]
\begin{center}
 \scalebox{1.0}{
     \begin{tabular}{|l|l|l|l|}
     \hline
     \textbf{Model / Parameters}                 & \textbf{min} & \textbf{max} & \textbf{mean} \\ \hline
     \textbf{Boids model}               & ~   & ~   & ~    \\
     radius ($m$)                & 0.1 & 1   & 0.3  \\
     comfort speed ($m/s$)      & 1   & 2   & 1.5  \\ \hline
     \textbf{Helbing model}             & ~   & ~   & ~    \\
     radius ($m$)                & 0.1 & 1   & 0.3  \\
     comfort speed ($m/s$)      & 1   & 2   & 1.5  \\ \hline
     \textbf{RVO model}                & ~   & ~   & ~    \\
     comfort speed ($m/s$)      & 1   & 2   & 1.5  \\
     neighbor distance ($m$)     & 2   & 20  & 11   \\
     radius ($m$)                & 0.2 & 0.8 & 0.5  \\
     agent time horizon ($s$)    & 0.1 & 5   & 2    \\
     obstacle time horizon ($s$) & 0.1 & 5   & 2    \\ \hline
    \end{tabular}}
 \caption{Initial motion model parameter for optimization.}

\end{center}
\end{table}

\begin{table*}[ht]
\begin{center}
\scalebox{1.0}{
    \begin{tabular}{|l|l|l|l|}
    \hline
		\textbf{Dataset} & \textbf{Challenges} & \textbf{Density} & \textbf{Agents}                                       \\ \hline
    NDLS-1 & BV, PO, IC & High  & 131                                     \\ \hline
    IITF-1    & BV, PO, IC, CO  & High & 167\\ \hline
    IITF-3    & BV, PO, IC, CO  & High & 189\\ \hline
    IITF-5    & BV, PO, IC, CO  & High & 71\\ \hline
    NPLC-1    & BV, PO, IC  & Medium   & 79             \\ \hline
		NPLC-3   & BV, PO, IC, CO & Medium & 144\\ \hline
    IITF-2    & BV, PO, IC, CO  & Medium & 68                \\ \hline
    \end{tabular}}
\scalebox{1.0}{
    \begin{tabular}{|l|l|l|l|}
    \hline
		\textbf{Dataset} & \textbf{Challenges} & \textbf{Density} & \textbf{Agents}                                       \\ \hline
    IITF-4    & BV, PO, IC, CO  & Medium &  116              \\ \hline
    NDLS-2   & BV, PO, IC, CO & Low & 72 \\ \hline
	  NPLC-2 & BV, PO & Low & 56                                       \\ \hline
		seq\_hotel & IC, PO & Low & 390                                       \\ \hline
		seq\_eth  & BV, IC, PO & Low & 360                                       \\ \hline
		zara01 & BV, IC, PO & Low & 148                                       \\ \hline
		zara02 & BV, IC, PO & Low & 204                                       \\ \hline
    \end{tabular}}
\caption{Crowd Scenes used as Benchmarks. We highlight many attributes of crowd these videos along with density and the number of number of pedestrians tracked. We use the following abbreviations about the underlying scene: Background Variations(BV), Partial Occlusion(PO), Complete Occlusion(CO), Illumination Changes(IC)}
\end{center}
\label{tb:tablediv}
\end{table*}

\begin{table*}[ht]
\resizebox{\textwidth}{!}{%
\begin{tabular}{|c|c|c|c|c|c|c|c|c|c|c|c|c|
>{\columncolor[HTML]{FFCCC9}}c |
>{\columncolor[HTML]{FFCCC9}}c |
>{\columncolor[HTML]{FFCCC9}}c |}
\hline
 & \multicolumn{3}{c|}{LIN} & \multicolumn{3}{c|}{Boids} & \multicolumn{3}{c|}{Helbing} & \multicolumn{3}{c|}{RVO} & \multicolumn{3}{c|}{\cellcolor[HTML]{FFCCC9}\textbf{MMM}} \\ \hline
 & LD & MD & HD & LD & MD & HD & LD & MD & HD & LD & MD & HD & \textbf{LD} & \textbf{MD} & \textbf{HD} \\ \hline
\textbf{MOTP} & 64.42\% & 52.82\% &  & 67.24\% & 57.10\% & 43.14\% & 70.52\% & 61.33\% & 49.88\% & 72.19\% & 63.17\% & 51.31\% & \textbf{73.98\%} & \textbf{69.23\%} & \textbf{54.29\%} \\ \hline
\textbf{MOTA} & 49.42\% & 35.3\% & 31.37\% & 50.59\% & 26.42\% & 40.88\% & 53.28\% & 44.19\% & 33.51\% & 53.95\% & 48.81\% & 35.83\% & \textbf{54.18\%} & \textbf{50.16\%} & \textbf{38.83\%} \\ \hline
\end{tabular}
}
\caption{
We compare the MOTA and MOTP values across the density groups and the different motion models.
}
\label{CLEAR}
\end{table*}

\begin{table*}[ht]
\centering
\resizebox{\textwidth}{!}{%

\begin{tabular}{|c|c|c|c|c|c|c|c|c|c|c|c|c|c|c|c|c|c|c|c|c|}

\hline

&
\multicolumn{8}{|c}{High Density} &
\multicolumn{8}{|c}{Medium Density} &
\multicolumn{4}{|c|}{Low Density} \\
\hline
         
&

\multicolumn{2}{|c}{\textbf{NDLS-1}} &
\multicolumn{2}{|c}{\textbf{IITF-1}} &
\multicolumn{2}{|c}{\textbf{IITF-3}} &
\multicolumn{2}{|c}{\textbf{IITF-5}} &
\multicolumn{2}{|c}{\textbf{NPLC-1}} &
\multicolumn{2}{|c}{\textbf{NPLC-3}} &
\multicolumn{2}{|c}{\textbf{IITF-2}} &
\multicolumn{2}{|c}{\textbf{IITF-4}} &
\multicolumn{2}{|c}{\textbf{NDLS-2}} &
\multicolumn{2}{|c|}{\textbf{NPLC-2}} \\
\cline{2-21}

 & ST & IS & ST & IS & ST & IS & ST & IS & ST & IS & ST & IS & ST & IS & ST & IS & ST& IS & ST & IS \\
 \hline
 
LIN & 53 & 17 & 63 & 27 & 51 & 35 & 59 & 18 & 67 & 15 & 60 & 29 & 36 & 22 & 52 & 36 & 68 & 23 & 69 & 21 \\
\hline

Boids & 58 & 15 & 66 & 23 & 56 & 33 & 65 & 14 & 73 & 13 & 65 & 26 & 40 & 19 & 52 & 35 & 70 & 22 & 72 & 19 \\
\hline

Helbing & 56 & 16 & 66 & 26 & 52 & 33 & 62 & 15 & 74 & 11 & 68 & 23 & 41 & 19 & 59 & 31 & 75 & 18 & 72 & 14 \\
\hline

LTA & 54 & 17 & 65 & 22 & 51 & 32 & 60 & 17 & 68 & 11 & 62 & 28 & 42 & 18 & 54 & 32 & 69 & 23 & 70 & 20 \\
\hline

RVO & 57 & 14 & 69 & 20 & 53 & 29 & 64 & 13 & 71 & 10 & 64 & 26 & 42 & 18 & 53 & 32 & 72 & 20 & 74 & 16 \\
\hline


MeanShift & 27 & 32 & 31 & 38 & 23 & 52 & 34 & 29 & 39 & 36 & 41 & 31 & 22 & 33 & 39 & 45 & 31 & 28 & 45 & 28 \\
\hline

\rowcolor[HTML]{FFCCC9}MMM & 63 & 12 & 73 & 19 & 57 & 27 & 67 & 10 & 77 & 7 & 71 & 20 & 44 & 16 & 63 & 28 & 79 & 17 & 78 & 14 \\
\hline

\end{tabular}

}
\caption{
We compare the percentage of successful tracks (ST) and ID switches (IS) of our mix motion model algorithm (MMM) with homogeneous motion models - LIN, Boids, Helbing, LTA, RVO and a baseline mean-shift tracker.
Our method provides higher accuracy compared to homogeneous motion models and lesser ID switches.
The benefits of our approach is higher, as the crowd density increases. These datasets are publicly available at http://gamma.cs.unc.edu/RCrowdT/.
}
\label{tb:tablescore}
\end{table*}

\begin{table*}[ht]
\centering
\scalebox{1.0}{

\begin{tabular}{|c|c|c|c|c|c|c|c|c|c|c|c|c|c|c|c|c|c|c|c|c|}

\hline

&

\multicolumn{2}{|c}{\textbf{seq\_hotel}} &
\multicolumn{2}{|c}{\textbf{seq\_eth}} &
\multicolumn{2}{|c}{\textbf{zara01}} &
\multicolumn{2}{|c|}{\textbf{zara02}}  \\
\cline{2-9}

 & ST & IS & ST & IS & ST & IS & ST & IS  \\
 \hline
 
LIN & 182 & 92 & 187 & 58 & 51 & 27 & 49 & 27  \\
\hline

Boids & 192 & 78 & 202 & 59 & 52 & 27 & 54 & 26  \\
\hline

Helbing & 221 & 73 & 232 & 48 & 54 & 26 & 55 & 25  \\
\hline

LTA & 238 & 70 & 249 & 42 & 60 & 24 & 62 & 25 \\
\hline

RVO & 241 & 71 & 258 & 37 & 61 & 22 & 65 & 23 \\
\hline


MeanShift & 98 & 171 & 112 & 139 & 32 & 41 & 33 & 39 \\
\hline

\rowcolor[HTML]{FFCCC9}MMM & 252 & 68 & 267 & 34 & 63 & 20 & 68 & 21 \\
\hline

\end{tabular}

}
\caption{We compare the percentage of successful tracks (ST) and ID switches (IS) of our mix motion model algorithm (MMM) with homogeneous motion models - LIN, Boids, Helbing, LTA, RVO and a baseline mean-shift tracker with standard datasets - seq\_hotel , seq\_eth , zara01 , zara02 ~\cite{pellegrini2010improving}.}
\label{tb:tablescore2}
\end{table*}

\begin{table*}[ht]
\centering
\scalebox{1.0}{

\begin{tabular}{|c|c|c|c|c|c|c|c|c|c|c|c|c|c|c|c|c|c|c|c|c|}

\hline

&

\multicolumn{8}{|c}{High Density} &
\multicolumn{6}{|c|}{Medium Density} \\
\hline

&

\multicolumn{2}{|c}{\textbf{NDLS-1}} &
\multicolumn{2}{|c}{\textbf{IITF-1}} &
\multicolumn{2}{|c}{\textbf{IITF-3}} &
\multicolumn{2}{|c}{\textbf{IITF-5}} &
\multicolumn{2}{|c}{\textbf{NPLC-1}} &
\multicolumn{2}{|c}{\textbf{NPLC-3}} &
\multicolumn{2}{|c|}{\textbf{IITF-2}}  \\
\cline{2-15}

 & ST & FPS & ST & FPS & ST & FPS & ST & FPS & ST & FPS & ST & FPS & ST & FPS  \\
 \hline
 
 MMM-C & 63 & 11 & 74 & 12 & 57 & 11 & 67 & 12 & 78 & 14 & 71 & 13 & 46 & 13 \\
\hline
 
\rowcolor[HTML]{FFCCC9} MMM & 63 & 27 & 73 & 28 & 57 & 26 & 67 & 26 & 77 & 28 & 71 & 26 & 44 & 26  \\
\hline

\end{tabular}
}

\vspace*{0.5 cm}

\scalebox{1.0}{

\begin{tabular}{|c|c|c|c|c|c|c|c|c|c|c|c|c|c|c|c|c|c|c|c|c|}

\hline

&

\multicolumn{2}{|c}{Medium Density} &
\multicolumn{12}{|c|}{Low Density}   \\
\hline

&

\multicolumn{2}{|c}{\textbf{IITF-4}} &
\multicolumn{2}{|c}{\textbf{NDLS-2}} &
\multicolumn{2}{|c|}{\textbf{NPLC-2}} &
\multicolumn{2}{|c}{\textbf{seq\_hotel}} &
\multicolumn{2}{|c}{\textbf{seq\_eth}} &
\multicolumn{2}{|c}{\textbf{zara01}} &
\multicolumn{2}{|c|}{\textbf{zara02}}   \\
\cline{2-15}

 & ST & FPS & ST & FPS & ST & FPS & ST & FPS & ST & FPS & ST & FPS & ST & FPS \\
 \hline

MMM-C & 63 & 11 & 80 & 12 & 78 & 11 & 254 & 11 & 267 & 16 & 63 & 14 & 69 & 15 \\
\hline

\rowcolor[HTML]{FFCCC9} MMM & 63 & 27 & 79 & 28 & 78 & 26 & 252 & 28 & 267 & 29 & 63 & 27 & 68 & 28  \\
\hline

\end{tabular}
}
\caption{
We compare the percentage of successful tracks (ST) and average tracking frames per second (FPS) of our mixture of motion models algorithm adaptive particle filtering (MMM) and with constant particle numbers (MMM-C).
}
\label{tb:tablefps}
\end{table*}

\begin{figure}[ht]
	\centering
		\includegraphics[width=0.4\textwidth]{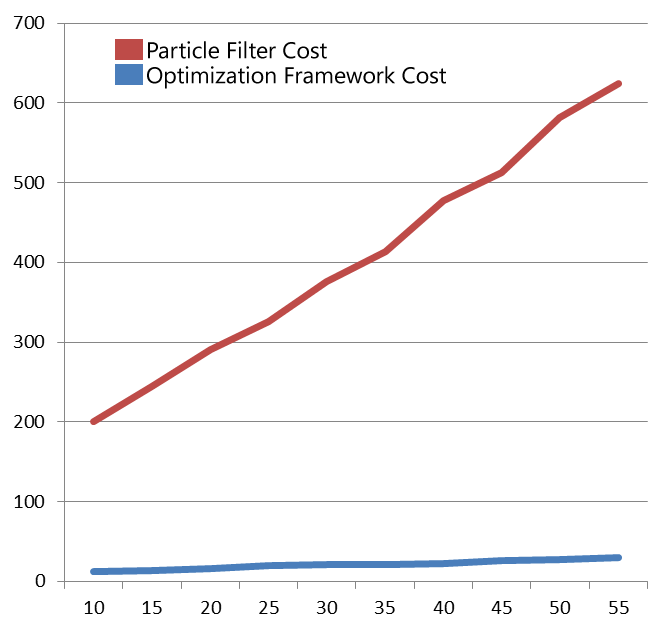}
		
\caption{
Computation cost comparison between the particle filter system and the optimization framework. The x-axis represents number of people tracked and the y-axis represent the computation time (in milliseconds) }
\label{fig:compcost}
\end{figure}

\section{Implementation and Results}\label{section:results}

In this section we present our implementation details and highlight the performance on 10 different crowd video datasets.

\subsection{Evaluation}
We use the \textbf{CLEAR MOT}~\cite{keni2008evaluating} evaluation metrics to analyze the performance analytically. We use the \textbf{MOTP} and the \textbf{MOTA} metrics. \textbf{MOTP} evaluates the alignment of tracks with the ground truth while \textbf{MOTA} produces a score based on the amount of false positives, missed detections, and identity switches. These metrics have become standard for evaluation of detection and tracking algorithms in the computer vision community, and we refer the interested reader to ~\cite{keni2008evaluating} for more a detailed explanation.

We analyze these metric across the density groups and the different motion models (Table~\ref{CLEAR}).

\subsection{Tracking Results}

We highlight the performance of our algorithm based on a mixture of motion models on different benchmarks, comparing the performance of our algorithm with single, homogeneous motion model methods: constant velocity model (LIN), LTA~\cite{pellegrini2009you}, Social Forces~\cite{yamaguchi2011you}, Boids~\cite{Reynolds1999} and RVO~\cite{van2011reciprocal}.
LIN models the velocities of pedestrians as constant, and is the underlying motion model frequently used in the standard particle filter.
The other four models compute the pedestrian states based on optimizing functions, which model collision avoidance, destinations of pedestrians, and the desired speed.
In our implementation, we replace the state transition process of a standard particle filtering algorithm with different motion models.

We evaluate on some challenging datasets~\cite{bera2014} which are available publicly and also some standard datasets from the pedestrian tracking community. These videos were recorded at 24-30 fps. We manually annotated these videos and corrected the perspective effect by camera calibration.
We also compare our performance compared to a baseline mean-shift tracker (Table~\ref{tb:tablescore}). We also compare the computational overhead of our optimization framework compared the particle filter system in terms of computation time. (Refer Figure~\ref{fig:compcost})

For our evaluation, we have divided our system into two phases:

\emph{Initialization:} Here we initialize the motion model estimation and parameter-optimization system with hand-drawn or ground truth data for a few initial frames, which is computed offline. For our experiments, we've used the first 10 frames. We compute a score that is used to choose the best-fit model from our motion model set and the associated parameters.

\emph{Prediction:} After learning from the initial data, we use the predicted set of parameters to model the state transition part of the
standard Bayesian inference framework. We iteratively and incrementally recompute the score and update the motion model. This computation is performed in realtime.

We show the number of correctly tracked pedestrians and the number of ID switches. A track is counted as ``successful'' when the estimated mean error between the tracking result and the ground-truth value is less than 0.8 meter in groundspace. The average human stride length is about 0.8 meter and we consider the tracking to be incorrect if the mean error is more than this value. 
Our method provides 9-18\% higher accuracy over LIN for medium density crowds (Table~\ref{tb:tablescore}).
Moreover, we compare the performance of our adaptive particle tracking algorithm with a particle filter that uses constant number of particles (Table~\ref{tb:tablefps}).

\begin{figure*}[!htb]
	\centering
		\includegraphics[width=1\textwidth]{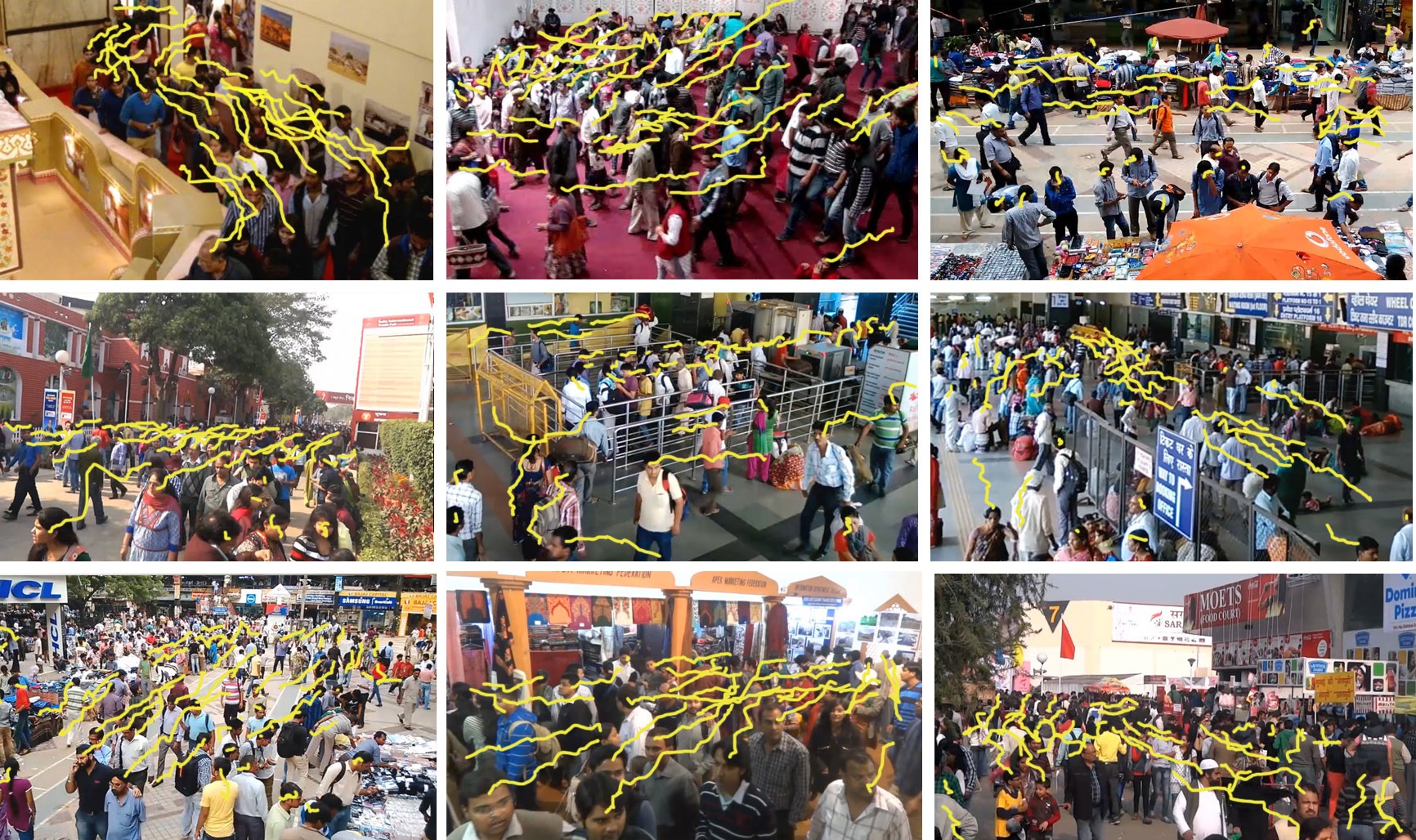}
		\label{fig:9}
\caption{
The results of our approach on some challenging datasets.
From top to bottom, left to right: IITF-1, IITF-2, NPLC-1, IITF-3, NDLS-2, NDLS-1, NLPC-2, IITF-4, IITF-5.
We are able to achieve a 4-12\% increase in accuracy over homogeneous motion models at interactive framerates.
}
\vspace*{-0.13in}
\end{figure*}

\section{Limitations, Conclusions, and Future Work}\label{section:conclusion}

We present a realtime algorithm for pedestrian tracking in crowded scenes.
Our algorithm provides a good balance between accuracy and speed.
We highlight its performance on many pedestrian datasets, showing that it can track crowded scenes in realtime on a PC with a multi-core CPU.
As compared to prior algorithms of similar accuracy, we obtain 2-3 times speedup.

Our approach has some limitations related to our motion model. Our motion model set does not take into account physiological and psychological pedestrian traits. All pedestrians are modeled with the same sensitivity towards gender and density; our model doesn't take into account heterogeneous agent characteristics, which affect the final behavior. These behavior characteristics can introduce additional errors in our confidence estimation. In practice, the performance of the algorithm can vary based on various other attributes of the input video.

As part of future work, we would like to incorporate the personality characteristics of the pedestrians, along with other characteristics, such as `fundamental diagrams' from pedestrian dynamics. We would like to parallelize the approach on a GPU to handle more complex pedestrian datasets in realtime. Finally, we would like to use improved learning algorithms to increase the accuracy of our tracker.


\section{Acknowledgements}\label{section:Acknowledgements}

This work was supported by NSF awards 1000579, 1117127, 1305286, Intel Corporation, and a grant from the Boeing Company


\bibliographystyle{spmpsci}
\bibliography{references}

\end{document}